\documentclass{article}
\usepackage{spconf,amsmath,amssymb,graphicx}
\usepackage{gensymb}
\usepackage{glossaries}
\usepackage{booktabs}
\usepackage{multirow}
\usepackage{bm} 
\usepackage{subfloat}
\usepackage{subcaption}
\usepackage{xcolor}
\usepackage{flushend}
\usepackage{hyperref}
\usepackage[maxbibnames=4,doi=false,style=ieee,isbn=false,url=false,eprint=false]{biblatex}
\usepackage{enumitem}
\usepackage{balance}
\usepackage{graphicx}
\usepackage{bbding}
\usepackage{fixltx2e} 
\usepackage[font=scriptsize,labelfont=bf]{caption} 
\usepackage{float}

\def\Usetp{\mathcal{U}^p_{\{t_0, W \}}}

\def\Usetf{\mathcal{U}^f_{\{t_1, H \}}}
\def\Xtrainp{\mathbf{X}^p_{\{t_0, W \}}}
\def\Xgt{\mathbf{X}^f_{\{t_1, H \}}}
\def\Xhat{\hat{\mathbf{X}}^f_{\{t_1, H \}}}

\def\U{\mathbf{U}}

\def\Cond{\operatorname{Cond}}
\def\MLP{\operatorname{MLP}}
\def\Cp{\mathbf{C}^{p}}
\def\Cf{\mathbf{C}^{f}}
\def\Ep{\mathbf{E}^{p}}
\def\Ef{\mathbf{E}^{f}}
\def\O{\mathbf{O}}

\newacronym{nn}{NN}{Neural Network}
\newacronym{gnn}{GNN}{Graph Neural Network}
\newacronym{stgnn}{STGNN}{Spatiotemporal Graph Neural Network}
\newacronym{gatedgnn}{GGNN}{Gated GNN}
\newacronym{fcgnn}{FC-GNN}{Fully Connected GNN}
\newacronym{gwavenet}{GWaveNet}{Graph WaveNet}
\newacronym{agcrn}{AGCRN}{Adaptive Graph Convolutional Recurrent Network}
\newacronym{magcrn}{MAGCRN}{Modified AGCRN}
\newacronym{mae}{MAE}{Mean Absolute Error}
\newacronym{rmse}{RMSE}{Root Mean Squared Error}
\newacronym{mre}{MRE}{Mean Relative Error}
\newacronym{napl}{NAPL}{Node Adaptive Parameter Learning}
\newacronym{dl}{DL}{Deep Learning}
\newacronym{aq}{AQ}{Air Quality}
\newacronym{sshgnn}{SSH-GNN}{Self-Supervised Hierarchical Graph Neural Network}
\newacronym{dcrnn}{DCRNN}{Diffusion Convolutional Recurrent Neural Network}
\newacronym{gcn}{GCN}{Graph Convolutional Network}
\newacronym{gc}{GC}{Graph Convolution}
\newacronym{lstm}{LSTM}{Long Short-Term Memory}
\newacronym{gru}{GRU}{Gated Recurrent Unit}
\newacronym{dgnaea}{DGN-AEA}{Dynamic Graph Neural
Network with Adaptive Edge Attributes}
\newacronym{mlp}{MLP}{Multilayer Perceptron}
\newacronym{fc}{FC}{Fully Connected}
\newacronym{relu}{ReLU}{Rectified Linear Unit}

\title{Back to the Future: GNN-based NO\textsubscript{2} Forecasting Via Future Covariates}
\name{\begin{tabular}{c}Antonio Giganti\(^1\), Sara Mandelli\(^1\), Paolo Bestagini\(^1\), \\ Umberto Giuriato\(^2\),  Alessandro D'Ausilio\(^2\), Marco Marcon\(^1\), Stefano Tubaro\(^1\)\end{tabular}}

\address{\(^1\)Dipartimento di Elettronica, Informazione e Bioingegneria, Politecnico di Milano - Milan, Italy \\
\(^2\)ARIANET srl, Via Benigno Crespi 52, Milan, Italy \\
\thanks{This work was supported by the Italian Ministry of University and
Research (MUR) and the European Union (EU) under the PON/REACT project.}}

\begin{document}
\ninept
\definecolor{blue}{HTML}{1f77b4}
\definecolor{orange}{HTML}{ff7f0e}
\definecolor{green}{HTML}{2ca02c}
\maketitle
\begin{abstract}
Due to the latest environmental concerns in keeping at bay contaminants emissions in urban areas, air pollution forecasting has been rising the forefront of all researchers around the world.
When predicting pollutant concentrations, it is common to 
include the effects of environmental factors that influence these concentrations within an extended period, like traffic, meteorological conditions and geographical information.
Most of the existing approaches exploit this information as \textit{past} covariates, i.e., past exogenous variables that affected the pollutant but were not affected by it. 
In this paper, we present a novel forecasting methodology to predict NO\textsubscript{2} concentration via both past and \textit{future} covariates. Future covariates are represented by weather forecasts and future calendar events, which are already known at prediction time. 
In particular, we deal with air quality observations in a city-wide network of ground monitoring stations, modeling the data structure and estimating the predictions with a \gls{stgnn}. 
We propose a conditioning block that embeds past and future covariates into the current observations. After extracting meaningful spatiotemporal representations, these are fused together and projected into the forecasting horizon to generate the final prediction.
To the best of our knowledge, it is the first time that future covariates are included in time series predictions in a structured way.
Remarkably, we find that conditioning on future weather information has a greater impact than considering past traffic conditions. 
We release our code implementation at \href{https://github.com/polimi-ispl/MAGCRN}{https://github.com/polimi-ispl/MAGCRN}.
\end{abstract}

\glsresetall


\section{Introduction}
\label{sec:intro}
\glsreset{gnn}

In the landscape of 21st-century environmental concerns, air pollution forecasting has risen to the forefront, capturing widespread attention at a global scale. Addressing this challenge, numerous researchers have delved into the exploration of effective solutions, endeavoring to precisely forecast air pollutant concentrations through a variety of methods. Among these approaches, \gls{dl} methods currently hold the predominant interest, marking significant advancements in the efforts to forecasting air quality parameters and mitigating the impact of air pollution~\cite{bendis_dl_ts_2022, jin_spatiotemporal_2023, brovelli_n02_2023}.

Pollutant concentrations exhibit intricate correlations in both temporal and spatial domains, and these correlations dynamically evolve over time~\cite{zhang_dl_pollutant_forecast_2022, jacob_climate_2009, liu_weather_pollution_2020}.
Recently, \gls{dl}-based approaches have demonstrated considerable success in pollutant forecasting by capturing nonlinear temporal and spatial patterns very effectively~\cite{jin_spatiotemporal_2023, demedrano_socaire_2021}.

In the last few years, there has been a surge in the popularity of applying \glspl{stgnn} to pollutants forecasting~\cite{han_semi_hierarchical_2023, jin_self_optimization_aqi_2023, xu_dynamic_2023, han_joint_2023, phung_unsupervised_aqi_2022, wang_inter-station_2021, yanlin_hybrid_2019} since they can process 
the data structure by modeling it as a graph.
Indeed, the underlying assumption is to incorporate the data structure as an inductive bias~\cite{alippi_gnn_forecasting_2023, longa_graph_2023, leus_graph_sp_2023}. 
Time-evolving observations coming from ground monitoring stations inherently contains rich spatiotemporal structure and spatiotemporal dynamics.
The dependency structure of the observations can be captured through pairwise relationships among the stations. These representations form a graph where each station corresponds to a node and the functional dependencies among the stations can be seen as edges.
Most of the existing approaches capture spatial dependencies on a fixed graph structure, assuming that the underlying relationship between entities is fixed and pre-determined~\cite{ditsuhi_madrid_2023, zhao_mastgn_2020, wang_domain_knowledge_2020, phung_unsupervised_aqi_2022}. However, the explicit graph structure may not necessarily reflect the true dependency between measurements and existing relationships may be missing due to incomplete data connections~\cite{zonghan_gwavenet_2019, alippi_gnn_survey_2023}.
Recently, graph learning methods have been proposed~\cite{alippi_sparse_learning_2023} that allow to learn the graph structure directly from data.
These techniques are promising, pushing forward the forecasting abilities~\cite{zonghan_gwavenet_2019, lei_agcrn_2020, wu_connecting_dots_2020}.

In the context of air pollutant forecasting like NO\textsubscript{2}, PM\textsubscript{10} and PM\textsubscript{2.5}, it is reasonable to assume that including auxiliary information helps the prediction performance. 
This is motivated by the fact that air quality is often influenced by exogenous variables like traffic and weather conditions. 
For example, traffic-related emissions have been one of the top contributors to air pollution in many cities around the world. It has been proved that these emissions can deteriorate ambient air quality on a large spatial scale, especially during the morning and evening rush hours in urban regions~\cite{zhang_dl_pollutant_forecast_2022, han_semi_hierarchical_2023, ditsuhi_madrid_2023}. 
In addition, studies have shown that air pollutants vary under different meteorological conditions~\cite{jacob_climate_2009, bertrand_cams_ml_2023}. 
Indeed, the temperature affects the atmospheric and ventilation conditions; humidity and precipitation can change the deposition characteristics of particulate matter; wind speed promotes the diffusion and spread of pollutants~\cite{jacob_introduction_1999}.

All this information could be included in the air quality forecasting as covariates, i.e., exogenous variables that affect the pollutant to predict, but are not affected by it.
For instance, past covariates are represented by traffic conditions, meteorological factors and geographical information. 
Future covariates are all the future-related information that are known in advance, like weather forecasts
and calendar events (seasons, days of a week, time of a day)~\cite{zhang_dl_pollutant_forecast_2022}.

Past covariates have been successfully exploited in~\cite{han_joint_2023, wang_inter-station_2021, ditsuhi_madrid_2023, phung_unsupervised_aqi_2022, yanlin_hybrid_2019, zhao_mastgn_2020}. 
To the best of our knowledge, very few \gls{stgnn}-based methods that exploit future covariates in performing pollutant prediction tasks have been proposed~\cite{han_semi_hierarchical_2023, xu_dynamic_2023, wang_domain_knowledge_2020}, but they lack a
structural way to consider future information into the forecasting process.

In this paper, we explore the potential of both past and \textit{future} covariates to perform forecasting of NO\textsubscript{2} concentration. 
To do so, we focus on \gls{stgnn}-based architectures that automatically learn the graph structure from data. 
We build our method upon the \gls{stgnn} module in~\cite{lei_agcrn_2020}, and we propose a conditioning block to incorporate additional past and future information into the forecasting process. Then, we condense the covariate information through a straightforward fusion process for predicting NO\textsubscript{2} concentration.

When tested on a recently released real-world air quality dataset, the proposed approach outperforms existing state-of-art forecasting methods, showing superior performances in predicting air quality parameters up to a forecasting horizon of three days. Interestingly, the role of future covariates reveals paramount for enhancing the prediction performances.  

\section{Problem Formulation}
\label{sec:problem}

We consider a collection of $N$, hourly-sampled, 1-dimensional time series, acquired from a network of $N$ ground monitoring stations. We denote with the matrix $\Xtrainp \in \mathbb{R}^{N\times W}$ the stacked $N$ observations acquired for $W$ consecutive \textit{past} time instants [hours] up to a final reference time instant $t_0$.
For instance, $\Xtrainp$ can represent the hourly evolution of NO\textsubscript{2} concentration acquired at $N$ different stations in the last $W$ hours.

The aim of this work is to perform a multi-step forecasting, that is, predicting for each time step in the forecasting horizon $H$ [hours] the \textit{future} observation $\Xgt \in \mathbb{R}^{N\times H}$ starting from a generic time instant $t_1 \geq t_0$ for a time window of $H$ hours. In general, $H$ can be different from $W$.
To face this problem, we exploit the knowledge of past observations $\Xtrainp$ combined with their covariate time series, i.e., the time-dependent exogenous variables that influence the time series in $\Xtrainp$  but are not affected by them.

We refer to the term \textit{past covariates} as the covariate time series known only into the past in the same time window of the acquired observations, i.e., in a temporal window $W$ up to the time instant $t_0$. An example of past covariate of NO\textsubscript{2} concentration is represented by the traffic data like the number of vehicles and their speed, which are strictly linked to air quality conditions.
Also past cyclical events derived from calendar like the daily time evolution and weekday information can be used as additional covariates~\cite{gardner_uk_daily_2000, stafoggia_estimation_2019}.
For simplicity, we denote with $\Usetp$ the set of time series associated with $P$ past covariates, being $\Usetp = \{ \U^p_i \in \mathbb{R}^{N \times W} \}, i\in [1, P]$. 

With the term \textit{future covariates} we refer to the covariate time series known into the future in the forecasting horizon, i.e.,
from the time instant $t_1$ for a time window $H$. This is the case for instance of future weather forecasts, which are known in advance and can provide useful information for predicting air quality conditions~\cite{jacob_climate_2009, jacob_introduction_1999}; future calendar events and daily/weekly time series can be used as well to predict air quality.
We define the set of time series related to $F$ future covariates as 
$\Usetf = \{ \U^f_i \in \mathbb{R}^{N \times H} \}, i\in [1, F]$. 

We formulate our forecasting problem as finding a function $F(\cdot)$ to forecast the next $H$ time steps observations based on the knowledge of the past temporal observations and past and future covariates. 
Therefore, 
\begin{equation*}
    \Xhat = F_{\theta}(\Xtrainp, \Usetp, \Usetf) 
\end{equation*}
where $\theta$ denotes all the learnable parameters of the forecasting model.


\section{Proposed Methodology}
\label{sec:method}

As recently done in the literature of time series forecasting~\cite{wu_dstcgcn_2023, wu_connecting_dots_2020, lei_agcrn_2020}, we model the tackled problem as a graph-based forecasting problem. 
Our network of ground monitoring stations can be represented with a graph structure, i.e., ground stations (graph nodes) that exhibit a certain degree of correlation (encoded in the graph edges). 

In particular, we propose a modified version of the \gls{agcrn} presented in~\cite{lei_agcrn_2020} and successfully used for predicting the future status of a traffic network from past data.
This network is part of a specific family of \glspl{gnn} known as \glspl{stgnn}, i.e., it is characterized by a graph topology that remains constant, while node and edge features change over time. 
We decided to get advantages of this architecture for two main reasons.
First, differently from others \glspl{stgnn}, \gls{agcrn} captures specific spatial and temporal characteristics of each node. This information improves the representation of station-level peculiarities and it is crucial since a station's measurement can greatly depend on its physical location (urban or rural)~\cite{alippi_local_effects_2023}.
Second,
the hidden graph topology is directly inferred from training data, 
which usually benefits the forecasting performance~\cite{alippi_gnn_forecasting_2023}.

\begin{figure}
    \centering
    \includegraphics[width=\columnwidth]{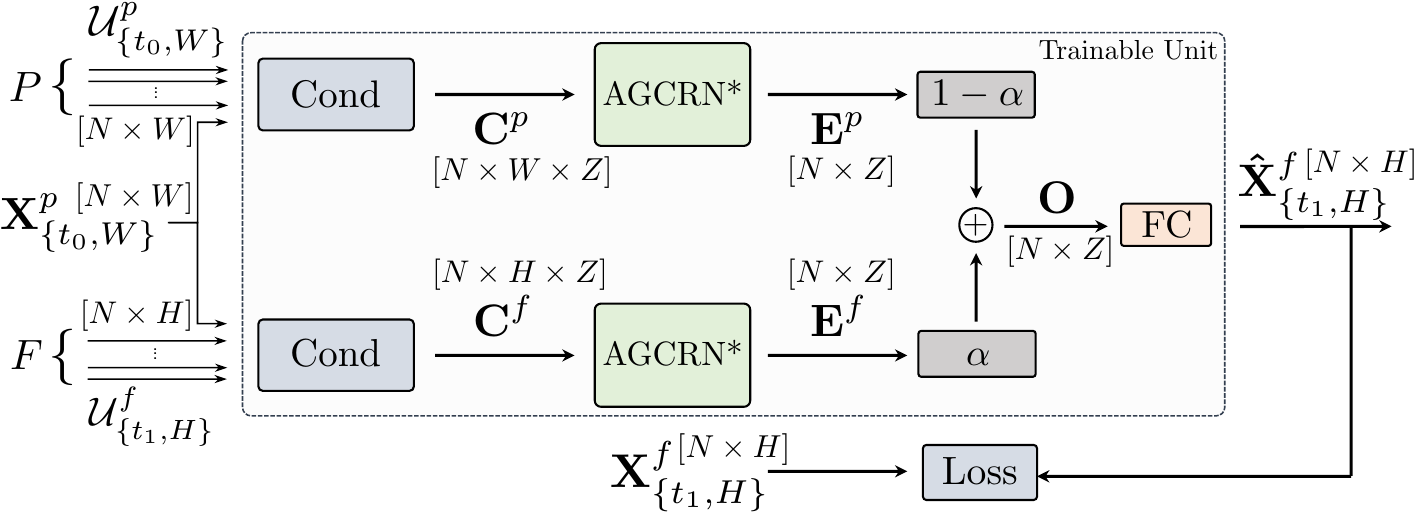}
    \caption{The training pipeline of the proposed MAGCRN. The AGCRN$^*$ used here represents AGCRN without the last FC layer.}
    \label{fig:system}
    \vspace{-12pt}
\end{figure}


\glsreset{magcrn}
To face our air quality forecasting task, we propose to modify the standard prediction paradigm of \gls{agcrn} by exploiting the knowledge of covariates.
Differently from the original version, our proposed \gls{magcrn} leverages the ability of 
conditioning the forecast not only on past but also on \textit{future} information. 
An example of future information that can influence our predictions is represented by the future weather forecasts~\cite{jacob_climate_2009, bertrand_cams_ml_2023}. 
By conditioning NO\textsubscript{2} forecasting on future weather forecasts, we can build a more comprehensive and accurate model that takes into account the intricate relationships between atmospheric conditions and air quality. 
It is worth noticing that, in a real scenario, future weather forecasts are available to the algorithm at prediction time.

The complete training pipeline of our proposed \gls{magcrn} is shown in Fig.~\ref{fig:system}.
We can recognize three main steps:
\begin{enumerate}
    \item \textit{Conditioning}, where the input observations $\Xtrainp$ are conditioned on past ($\Usetp$) and future ($\Usetf$) covariates separately, returning $\Cp$ and $\Cf$ signals. 
    \item \textit{Embedding estimation}, where we extract two embedding signals $\Ep$ and $\Ef$ from the conditioned signals $\Cp$ and $\Cf$. 
    \item \textit{Past \& Future fusion}, where we combine the $\Ep$ and $\Ef$ embeddings to construct the final 
    prediction
    $\Xhat$. 
\end{enumerate}

\noindent \textbf{Conditioning. }
The most important part of our system is the fully-learnable input conditioning module named $\Cond$, which conditions the current observed data $\Xtrainp$ on both past and future covariates. 
Given two generic data matrices $\mathbf{K}_1$ and $\mathbf{K}_2$, the proposed conditioning module takes inspiration from~\cite{torch_spatiotemporal_2022}, and can be computed as
\begin{equation*}
    \Cond(\mathbf{K}_1, \mathbf{K}_2) = \MLP(\phi(\MLP(\mathbf{K}_1))) + \MLP(\phi(\MLP(\mathbf{K}_2)))
\end{equation*}
where the $\MLP$ module is based on
a \gls{mlp} performing a linear transformation on the input data and $\phi$ is a non-linearity carried out by a \gls{relu}.
On the one hand, the input data are conditioned on past covariates, obtaining $\Cp= \Cond(\Xtrainp, \Usetp)$;  on the other hand, the same input is conditioned on future covariates, obtaining $\Cf=\Cond(\Xtrainp, \Usetf)$. 
In particular, $\Cp \in \mathbb{R}^{N\times W \times Z}$ and $\Cf \in \mathbb{R}^{N\times H \times Z}$, where $Z$ is the output dimension introduced by the \gls{mlp}-based conditioning modules. 
The structure of the two $\Cond$ blocks is the same; however, the four involved $\MLP$ blocks are different in the sense that their parameters are learned independently from each other.
\begin{figure}
    \centering
    \includegraphics[width=0.9\columnwidth]{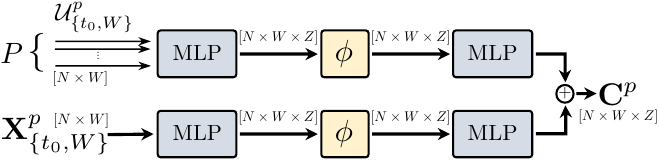}
    \caption{The proposed $\Cond$ module.}
    \label{fig:conditioning}
    \vspace{-12pt}
\end{figure}
In Fig.~\ref{fig:conditioning}, we report an example of the $\Cond$ module in conditioning the input window $\Xtrainp$ with past information $\Usetp$.

\noindent \textbf{Embedding estimation. }
After the conditioning, $\Cp$ and $\Cf$ are given as input to two identical \gls{agcrn} cells
to automatically learn the latent graph structure and, at the same time, capture node-specific patterns. 
To perform a late fusion between past and future conditioning contributions, these cells are modified by removing the last \gls{fc} layer. We define the outputs of the \gls{agcrn} cells as past and future embeddings, $\Ep$ and $\Ef$, respectively, with $\Ep, \Ef \in \mathbb{R}^{N\times Z}$. Notice that the temporal dimension has been embedded into the last features' dimension. 


\noindent \textbf{Past \& Future fusion. }
The embeddings are then fused together to obtain the final feature tensor $\O \in \mathbb{R}^{N\times Z}$, defined as
\begin{equation*}
    \O = (1-\alpha)\Ep + \alpha \Ef    
\end{equation*}
where $\alpha$ is a mixing coefficient determining the balance between the past and future information.
If $\alpha=0$, the output is entirely determined by past conditioning; if $\alpha=1$ the output relies only on future conditioning.
The resulting $\O$ vector is
mapped into the forecasting horizon $H$ by a linear \gls{fc} layer, obtaining an estimation of the future NO\textsubscript{2} concentration $\Xhat \in \mathbb{R}^{N\times H}$.

The network is trained minimizing the \gls{mae} between the groundtruth $\Xgt$ and its forecast $\Xhat$.

In testing phase, given $\Xtrainp$ along with its related past ($\Usetp$) and future ($\Usetf$) covariates, we exploit the trained \gls{magcrn} to perform the forecast, obtaining $\Xhat$.



\section{Experimental Setup}
\label{sec:setup}

\begin{figure}
    \centering
    \includegraphics[width=0.9\columnwidth]{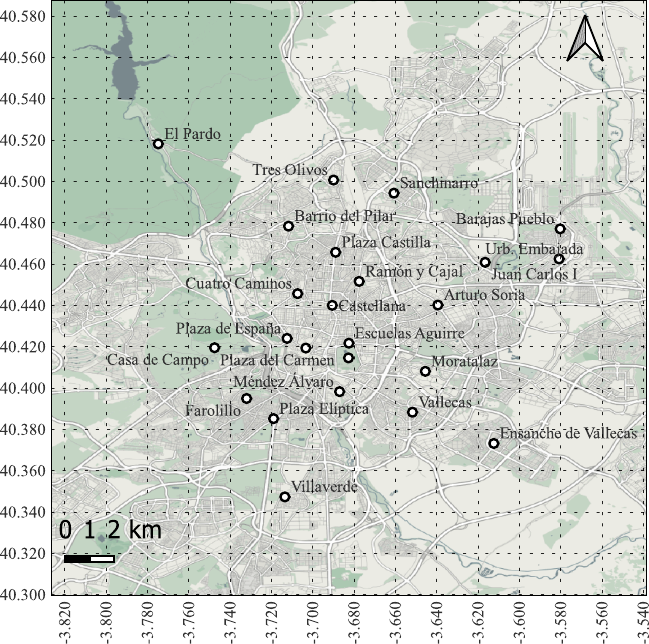}
    \caption{The distribution of the 24 ground monitoring stations placed in the city of Madrid. Each point represents one observation station.}
    \label{fig:station network}
    \vspace{-12pt}
\end{figure}

\noindent \textbf{Dataset. }We conduct experiments on a recently released real-world dataset presented in~\cite{ditsuhi_madrid_2023}. This dataset is composed of hourly-sampled air quality, meteorological and traffic data from 24 different ground monitoring stations in the city of Madrid (Spain), from January to June 2019. The station spatial distribution is reported in Fig.~\ref{fig:station network}. The released data include:
\begin{itemize}
    \item \textit{Air quality data} — NO\textsubscript{2} concentration $[\mu$g/m$^3]$;
    \item \textit{Meteorological data} — wind speed [m/s], wind direction [rad], temperature [\degree C], relative humidity [\%], barometric pressure [mb], solar irradiance [W/m$^2$];
    \item \textit{Traffic data} — intensity [vehicles/h], occupancy time [\%], load (degree of congestion) [\%], average traffic speed [km/h].
\end{itemize}

Our target data consists in the NO\textsubscript{2} concentration.
We consider as a past covariates the measurements related to the traffic conditions. 
Indeed, this information has been proven to be strongly correlated with air quality variations~\cite{zhang_dl_pollutant_forecast_2022, zhang_assembler_2015}.
As for future covariates, we consider all the meteorological data related to the forecasting horizon.
In addition, calendar-derived covariates
are included in both past a future covariates set since they are known in advance~\cite{zhang_dl_pollutant_forecast_2022}.

\vspace{1mm}
\noindent \textbf{Network baselines. } In this work, we focus on \glspl{stgnn} that directly learn the adjacency matrix of the graph nodes. 
To compare our proposed methodology with the state-of-the-art, we select three recently proposed \glspl{stgnn} for time series forecasting: 
\begin{itemize}
    \item \textit{Graph WaveNet}~\cite{zonghan_gwavenet_2019}, which integrates diffusion graph convolutions with dilated convolutions to capture spatial and temporal dependencies;
    \item \textit{\gls{gatedgnn}}~\cite{satorras_multivariate_2022},
    that integrates gated mechanisms and residual connections for effective information propagation;
    \item \textit{\gls{agcrn}}~\cite{lei_agcrn_2020}, the reference architecture which our method is built upon.
\end{itemize}
Notice that none of these \glspl{stgnn} directly supports future covariates, contrarily to our proposed methodology.

Moreover, we do not compare ourselves with state-of-the-art models exploiting future covariates since they address the problem in a different manner. The work in~\cite{han_semi_hierarchical_2023} adopts a hierarchical \gls{gnn} by constructing a three-level hierarchy, from cities to regions. Our data do not allow for this granularity as they regard a more limited area (city-wide). Similarly, in~\cite{xu_dynamic_2023} the authors focus on regional-level predictions. 

\vspace{1mm}
\noindent \textbf{Training and testing specifications. }
To train and test the selected \glspl{stgnn}, we split the dataset in chronological order by the ratio of 7:2:1 to generate training, validation, and test data.

To train our model, we construct consecutive input/output windows pairs with $W=24$ past information and $H=24$ future information, resulting in 3820
window pairs. 
We train our model in predicting 1-day ahead NO\textsubscript{2} concentration, thus we consider $t_1 = t_0$.
The employed \gls{mlp} in the $\Cond$ module consists in a \gls{fc} linear layer, with $Z = 64$.
Moreover, we equally weight past and future covariates' contributions, setting $\alpha = 0.5$. 
We combine the ADAM optimizer with the cosine annealing learning rate scheduler, setting the maximum and minimum learning rate to 10$^{-2}$ and 10$^{-7}$ respectively, with a maximum period of 20. We stop the learning phase if the validation error does not decrease after 100 epochs.

At testing stage we always evaluate a forecasting horizon $H=24$, considering 429 window pairs. However, in addition to the 1-day ($t_1 = t_0$) forecasting horizon, we also evaluate the model in performing 2-day ($t_1 = t_0 + 24$) and 3-day ($t_1 = t_0 + 48$) hourly NO\textsubscript{2} predictions without retraining the network. 
We do this for testing the robustness and the generalization capabilities of the proposed method to predict farther events in the future.
Then, we evaluate the model performance in terms of \gls{mae}, \gls{rmse} and \gls{mre}, i.e., the \gls{mae} normalized by the $\ell_1$-norm of the target window.

\section{Experimental Results}
\label{sec:results}

\begin{table}[]
\centering
\caption{Forecasting error by means of \gls{mae}, \gls{rmse}, \gls{mre}, considering the baseline networks and our proposed \gls{magcrn}. We can exploit two different groups of covariates (i.e., past $\mathcal{U}^p$ and future $\mathcal{U}^f$) for three different forecasting horizons, i.e., 1-24h (1-day),  25-48h (2-day), 49-72h (3-day). Best results in bold. }
\label{tab:results}
\resizebox{\columnwidth}{!}{%
\begin{tabular}{@{}lcclll@{}}
\toprule
\multicolumn{1}{c}{\multirow{2}{*}{Model}} &  &  & \multicolumn{3}{c}{Forecasting Error (1-day / 2-day / 3-day)} \\ \cmidrule(l){2-6} 
\multicolumn{1}{c}{} & $\Usetp$ & $\Usetf$ & \multicolumn{1}{c}{MAE} & \multicolumn{1}{c}{RMSE} & \multicolumn{1}{c}{MRE} \\ \midrule
Graph WaveNet & \Checkmark &  & 12.33 / 13.51 / 14.16 & 19.41 / 20.59 / 21.37 & 0.48 / 0.52 / 0.55 \\
\gls{gatedgnn} & \Checkmark &  & 11.53 / 13.93 / 14.01 & 18.15 / 20.75 / 20.74 & 0.44 / 0.54 / 0.54 \\
\gls{agcrn} & \Checkmark &  & 11.56 / 13.17 / 13.69 & 18.31 / 20.14 / 20.70 & 0.45 / 0.51 / 0.53 \\
\textbf{MAGCRN (ours)} & \Checkmark & \Checkmark & \textbf{9.43 / 10.30 / 10.85} & \textbf{14.64 / 15.56 / 16.19} & \textbf{0.36 / 0.40 / 0.42} \\ \bottomrule
\end{tabular}%
}
\vspace{-14pt}
\end{table}

\noindent \textbf{General results. } Tab.~\ref{tab:results} reports the overall results of \gls{magcrn} and the compared baselines in terms of \gls{mae}, \gls{rmse} and \gls{mre} for three different forecasting horizons.  As expected, the performances decrease as the forecasting horizon moves far from the 1-day predictions. Nonetheless, our approach, which jointly considers past and future covariates in learning the graph structure and performing the desired forecast, consistently outperforms all the baselines. 

Since NO\textsubscript{2} and air pollutants in general strongly depends on weather conditions~\cite{jacob_climate_2009, liu_weather_pollution_2020}, we believe that considering the future weather forecast as an additional conditioning information benefits the prediction. 
The combination of two different conditioning (past and future) brings more accurate predictions, showing the inherent dependence that exists between the pollutant and weather conditions.





\begin{figure}[t]
  \centering  
  \includegraphics[width=\columnwidth]{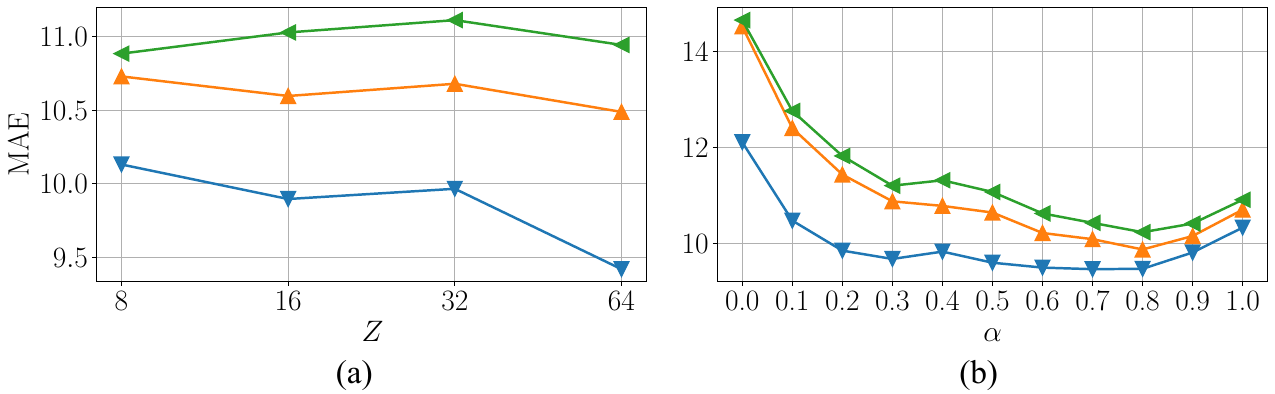}
  \caption{\gls{mae} results vs network parameters, according to different time forecasting horizons. In particular, \textcolor{blue}{$\blacktriangledown $} indicates 1-day time horizon, \textcolor{orange}{$\blacktriangle  $} the 2-day time horizon, \textcolor{green}{$\blacktriangleleft  $} the 3-day time horizon. }
  \label{fig:mae_parameter_sensitivity}
  \vspace{-14pt}
\end{figure}

\vspace{1mm}
\noindent \textbf{Parameter Sensitivity. } In this section, we investigate the sensitivity of the most important parameters of the proposed \gls{magcrn} for our forecasting task. 
Specifically, we explore the role of: (i) the embedding size $Z$; 
(ii) the $\alpha$ factor used to balance the action between past and future conditioning.
For the sake of brevity, we only report the achieved \gls{mae} as a function of the analyzed parameters, for different forecasting horizons. Each time we vary a parameter, we set the other to its default value, i.e., $Z=64$ and $\alpha=0.5$.
The results are reported in Fig.~\ref{fig:mae_parameter_sensitivity}.

Fig.~\ref{fig:mae_parameter_sensitivity} (a) depicts the performances by varying $Z$, that directly influences the generation of node-specific parameters in the \gls{stgnn}.
This parameter proves crucial for the near future, i.e., in the 1-day forecasting horizon, with an optimal value of $64$. On the contrary, $Z$ does not seem to impact more distant forecasting horizons (especially the 3-day case), since the trend is approximately stable for all the tested values.

In Fig.~\ref{fig:mae_parameter_sensitivity} (b) we vary the $\alpha$ factor used to balancing the action between the past and future conditioning. 
A lower $\alpha$ promotes the conditioning of past covariates in generating the forecast, up to the extreme case where the model considers only past covariates $\Usetp$ , i.e., $\alpha=0$.
On the other hand, a higher $\alpha$ favors the action of the future conditioning, up to the extreme case in which $\alpha=1$, where we discard $\Usetp$ and consider only $\Usetf$ in generating the forecast. 
The value $\alpha=0.5$ gives equal importance to past and future covariates in the prediction process.

It is possible to notice a particular behavior, consistent among all the analyzed forecasting horizons: assigning equal importance to past and future information
does not lead to the best results. 
The weather forecasts seem to play a paramount role in the NO\textsubscript{2} prediction. Indeed, the best performance corresponds to giving more importance ($\alpha=0.8$) to the future weather conditions rather than to the past information, which is provided by the traffic conditions. 
The extreme case in which $\alpha=1$, i.e., considering only the weather forecasts as covariates, does not bring the best results. 
Nonetheless, for the forecasting horizons that are distant to the one adopted in training (2-day and 3-day scenarios), the future conditioning becomes even more crucial to achieve acceptable results. 

\section{Conclusions}
\label{sec:conclusion}
\glsreset{stgnn}

This paper introduces an innovative forecasting methodology for predicting NO\textsubscript{2} time series concentrations by effectively leveraging past and future covariates. 
Differently from the solutions proposed in the literature, this is the first instance of future covariates being systematically included in time series predictions. 

The proposed study addresses air quality observations in a city-wide network of ground monitoring stations, utilizing a \gls{stgnn} to model the data structure and make predictions.
We incorporate past covariates (e.g., traffic conditions) and future covariates (e.g., weather forecasts and upcoming calendar events) through an input conditioning module. 
Then, we extract spatiotemporal representations of NO\textsubscript{2} observations conditioned to past and future covariates. We integrate these representations and project them into the forecasting horizon for the final forecast. 
Our findings emphasize the importance of including future information, especially for longer forecasting horizons, surpassing the impact of past traffic conditions.

Future work will delve more deeply into the role of the coefficient balancing past and future covariates contributions, enabling it to be learned in the training phase 
for enhanced performances.
Moreover, we will investigate potential improvements to the network architecture, integrating future covariates into the network's message-passing process and exploring dynamic graph structures.

\section{References}
\AtNextBibliography{\fontsize{8}{8}\selectfont}
\printbibliography[heading=none]

@misc{satorras_multivariate_2022,
	title = {Multivariate {Time} {Series} {Forecasting} with {Latent} {Graph} {Inference}},
	url = {http://arxiv.org/abs/2203.03423},
	doi = {10.48550/arXiv.2203.03423},
	urldate = {2024-01-03},
	publisher = {arXiv},
	author = {Satorras, Victor Garcia and Rangapuram, Syama Sundar and Januschowski, Tim},
	year = {2022},
	note = {arXiv:2203.03423 [cs]},
}

@inproceedings{zonghan_gwavenet_2019,
author = {Wu, Zonghan and Pan, Shirui and Long, Guodong and Jiang, Jing and Zhang, Chengqi},
title = {Graph Wavenet for Deep Spatial-Temporal Graph Modeling},
year = {2019},
booktitle = {Proceedings of the 28th International Joint Conference on Artificial Intelligence},
}

@inproceedings{lei_agcrn_2020,
author = {Bai, Lei and Yao, Lina and Li, Can and Wang, Xianzhi and Wang, Can},
title = {Adaptive Graph Convolutional Recurrent Network for Traffic Forecasting},
year = {2020},
isbn = {9781713829546},
publisher = {Curran Associates Inc.},
address = {Red Hook, NY, USA},
booktitle = {Proceedings of the 34th International Conference on Neural Information Processing Systems},
articleno = {1494},
numpages = {12},
location = {Vancouver, BC, Canada},
series = {NeurIPS'20}
}

@article{ditsuhi_madrid_2023,
  author={Iskandaryan, Ditsuhi and Ramos, Francisco and Trilles, Sergio},
  journal={IEEE Access}, 
  title={Graph Neural Network for Air Quality Prediction: A Case Study in Madrid}, 
  year={2023},
  volume={11},
  number={},
  pages={2729-2742},
  doi={10.1109/ACCESS.2023.3234214}}

@article{jacob_climate_2009,
title = {Effect of climate change on air quality},
journal = {Atmospheric Environment},
volume = {43},
number = {1},
pages = {51-63},
year = {2009},
note = {Atmospheric Environment},
issn = {1352-2310},
doi = {https://doi.org/10.1016/j.atmosenv.2008.09.051},
url = {https://www.sciencedirect.com/science/article/pii/S1352231008008571},
author = {Daniel J. Jacob and Darrell A. Winner},
}

@article{liu_weather_pollution_2020,
author = {Liu, Yansui and Zhou, Yang and Lu, Jiaxin},
year = {2020},
pages = {14518},
title = {Exploring the relationship between air pollution and meteorological conditions in China under environmental governance},
volume = {10},
journal = {Scientific Reports},
doi = {10.1038/s41598-020-71338-7}
}

@inproceedings{zhang_assembler_2015,
author = {Zhang, Chao and Zheng, Yu and Ma, Xiuli and Han, Jiawei},
title = {Assembler: Efficient Discovery of Spatial Co-Evolving Patterns in Massive Geo-Sensory Data},
year = {2015},
isbn = {9781450336642},
publisher = {Association for Computing Machinery},
address = {New York, NY, USA},
url = {https://doi.org/10.1145/2783258.2783394},
doi = {10.1145/2783258.2783394},
booktitle = {Proceedings of the 21th ACM SIGKDD International Conference on Knowledge Discovery and Data Mining},
pages = {1415–1424},
numpages = {10},
location = {Sydney, NSW, Australia},
series = {KDD '15}
}

@article{zhang_dl_pollutant_forecast_2022,
title = {Deep learning for air pollutant concentration prediction: A review},
journal = {Atmospheric Environment},
volume = {290},
pages = {119347},
year = {2022},
issn = {1352-2310},
doi = {https://doi.org/10.1016/j.atmosenv.2022.119347},
url = {https://www.sciencedirect.com/science/article/pii/S1352231022004125},
author = {Bo Zhang and Yi Rong and Ruihan Yong and Dongming Qin and Maozhen Li and Guojian Zou and Jianguo Pan},
}

@misc{alippi_gnn_survey_2023,
      title={A Survey on Graph Neural Networks for Time Series: Forecasting, Classification, Imputation, and Anomaly Detection}, 
      author={Ming Jin and Huan Yee Koh and Qingsong Wen and Daniele Zambon and Cesare Alippi and Geoffrey I. Webb and Irwin King and Shirui Pan},
      year={2023},
      eprint={2307.03759},
      archivePrefix={arXiv},
      primaryClass={cs.LG}
}

@misc{alippi_gnn_forecasting_2023,
      title={Graph Deep Learning for Time Series Forecasting}, 
      author={Andrea Cini and Ivan Marisca and Daniele Zambon and Cesare Alippi},
      year={2023},
      eprint={2310.15978},
      archivePrefix={arXiv},
      primaryClass={cs.LG}
}

@article{alippi_sparse_learning_2023,
  author  = {Andrea Cini and Daniele Zambon and Cesare Alippi},
  title   = {Sparse Graph Learning from Spatiotemporal Time Series},
  journal = {Journal of Machine Learning Research},
  year    = {2023},
  volume  = {24},
  number  = {242},
  pages   = {1--36},
  url     = {http://jmlr.org/papers/v24/22-1154.html}
}

@inproceedings{alippi_local_effects_2023,
author = {Andrea Cini and Ivan Marisca and Daniele Zambon and Cesare Alippi},
title = {Taming Local Effects in Graph-based Spatiotemporal Forecasting},
year = {2023},
booktitle = {Proceedings of the 37th International Conference on Neural Information Processing Systems},
location = {New Orleans, USA},
series = {NeurIPS'23}
}

@article{han_semi_hierarchical_2023,
  author={Han, Jindong and Liu, Hao and Xiong, Haoyi and Yang, Jing},
  journal={IEEE Transactions on Knowledge and Data Engineering}, 
  title={Semi-Supervised Air Quality Forecasting via Self-Supervised Hierarchical Graph Neural Network}, 
  year={2023},
  volume={35},
  number={5},
  pages={5230-5243},
  doi={10.1109/TKDE.2022.3149815}}

@inproceedings{phung_unsupervised_aqi_2022,
author = {Phung, Thu Hang and Nguyen, Duc Long and Vu, Viet Hung and Huynh, Thanh Trung and Nguyen, Thanh Hung and Nguyen, Phi Le},
title = {Unsupervised Air Quality Interpolation with Attentive Graph Neural Network},
year = {2022},
isbn = {9781450397254},
publisher = {Association for Computing Machinery},
address = {New York, NY, USA},
url = {https://doi.org/10.1145/3568562.3568657},
doi = {10.1145/3568562.3568657},
booktitle = {Proceedings of the 11th International Symposium on Information and Communication Technology},
pages = {103–110},
numpages = {8},
location = {Hanoi, Vietnam},
series = {SoICT '22}
}

@inproceedings{zhao_mastgn_2020,
  author={Zhao, Peijiang and Zettsu, Koji},
  booktitle={2020 IEEE International Conference on Big Data (Big Data)}, 
  title={MASTGN: Multi-Attention Spatio-Temporal Graph Networks for Air Pollution Prediction}, 
  year={2020},
  volume={},
  number={},
  pages={1442-1448},
  doi={10.1109/BigData50022.2020.9378156}}

@article{jin_self_optimization_aqi_2023,
AUTHOR = {Jin, Xue-Bo and Wang, Zhong-Yao and Kong, Jian-Lei and Bai, Yu-Ting and Su, Ting-Li and Ma, Hui-Jun and Chakrabarti, Prasun},
TITLE = {Deep Spatio-Temporal Graph Network with Self-Optimization for Air Quality Prediction},
JOURNAL = {Entropy},
VOLUME = {25},
YEAR = {2023},
NUMBER = {2},
ARTICLE-NUMBER = {247},
URL = {https://www.mdpi.com/1099-4300/25/2/247},
PubMedID = {36832613},
ISSN = {1099-4300},
DOI = {10.3390/e25020247}
}

@misc{wu_dstcgcn_2023,
      title={DSTCGCN: Learning Dynamic Spatial-Temporal Cross Dependencies for Traffic Forecasting}, 
      author={Binqing Wu and Ling Chen},
      year={2023},
      eprint={2307.00518},
      archivePrefix={arXiv},
      primaryClass={cs.LG}
}

@inproceedings{wu_connecting_dots_2020,
author = {Wu, Zonghan and Pan, Shirui and Long, Guodong and Jiang, Jing and Chang, Xiaojun and Zhang, Chengqi},
title = {Connecting the Dots: Multivariate Time Series Forecasting with Graph Neural Networks},
year = {2020},
isbn = {9781450379984},
publisher = {Association for Computing Machinery},
address = {New York, NY, USA},
url = {https://doi.org/10.1145/3394486.3403118},
doi = {10.1145/3394486.3403118},
booktitle = {Proceedings of the 26th ACM SIGKDD International Conference on Knowledge Discovery \& Data Mining},
pages = {753–763},
numpages = {11},
location = {Virtual Event, CA, USA},
series = {KDD '20}
}

@article{bendis_dl_ts_2022,
author = {Benidis, Konstantinos and Rangapuram, Syama Sundar and Flunkert, Valentin and Wang, Yuyang and Maddix, Danielle and Turkmen, Caner and Gasthaus, Jan and Bohlke-Schneider, Michael and Salinas, David and Stella, Lorenzo and Aubet, Fran\c{c}ois-Xavier and Callot, Laurent and Januschowski, Tim},
title = {Deep Learning for Time Series Forecasting: Tutorial and Literature Survey},
year = {2022},
issue_date = {June 2023},
publisher = {Association for Computing Machinery},
address = {New York, NY, USA},
volume = {55},
number = {6},
issn = {0360-0300},
url = {https://doi.org/10.1145/3533382},
doi = {10.1145/3533382},
journal = {ACM Comput. Surv.},
articleno = {121},
numpages = {36}
}

@misc{jin_spatiotemporal_2023,
      title={Spatio-Temporal Graph Neural Networks for Predictive Learning in Urban Computing: A Survey}, 
      author={Guangyin Jin and Yuxuan Liang and Yuchen Fang and Zezhi Shao and Jincai Huang and Junbo Zhang and Yu Zheng},
      year={2023},
      eprint={2303.14483},
      archivePrefix={arXiv},
      primaryClass={cs.LG}
}

@misc{longa_graph_2023,
      title={Graph Neural Networks for temporal graphs: State of the art, open challenges, and opportunities}, 
      author={Antonio Longa and Veronica Lachi and Gabriele Santin and Monica Bianchini and Bruno Lepri and Pietro Lio and Franco Scarselli and Andrea Passerini},
      year={2023},
      eprint={2302.01018},
      archivePrefix={arXiv},
      primaryClass={cs.LG}
}

@article{leus_graph_sp_2023,
  author={Leus, Geert and Marques, Antonio G. and Moura, José M.F. and Ortega, Antonio and Shuman, David I},
  journal={IEEE Signal Processing Magazine}, 
  title={Graph Signal Processing: History, development, impact, and outlook}, 
  year={2023},
  volume={40},
  number={4},
  pages={49-60},
  doi={10.1109/MSP.2023.3262906}}

@article{xu_dynamic_2023,
	title = {Dynamic graph neural network with adaptive edge attributes for air quality prediction: {A} case study in {China}},
	volume = {9},
	issn = {24058440},
	shorttitle = {Dynamic graph neural network with adaptive edge attributes for air quality prediction},
	url = {https://linkinghub.elsevier.com/retrieve/pii/S240584402304954X},
	doi = {10.1016/j.heliyon.2023.e17746},
	language = {en},
	number = {7},
	journal = {Heliyon},
	author = {Xu, Jing and Wang, Shuo and Ying, Na and Xiao, Xiao and Zhang, Jiang and Jin, Zhiling and Cheng, Yun and Zhang, Gangfeng},
	year = {2023},
}

@software{torch_spatiotemporal_2022,
    author = {Cini, Andrea and Marisca, Ivan},
    license = {MIT},
    month = {3},
    title = {{Torch Spatiotemporal}},
    url = {https://github.com/TorchSpatiotemporal/tsl},
    year = {2022}
}

@article{stafoggia_estimation_2019,
	title = {Estimation of daily {PM10} and {PM2}.5 concentrations in {Italy}, 2013–2015, using a spatiotemporal land-use random-forest model},
	volume = {124},
	issn = {0160-4120},
	url = {https://www.sciencedirect.com/science/article/pii/S0160412018327685},
	doi = {10.1016/j.envint.2019.01.016},
	language = {en},
	urldate = {2023-05-26},
	journal = {Environment International},
	author = {Stafoggia, Massimo and Bellander, Tom and Bucci, Simone and Davoli, Marina and de Hoogh, Kees and de' Donato, Francesca and Gariazzo, Claudio and Lyapustin, Alexei and Michelozzi, Paola and Renzi, Matteo and Scortichini, Matteo and Shtein, Alexandra and Viegi, Giovanni and Kloog, Itai and Schwartz, Joel},
	year = {2019},
	pages = {170--179},
}

@article{gardner_uk_daily_2000,
title = {Meteorologically adjusted trends in UK daily maximum surface ozone concentrations},
journal = {Atmospheric Environment},
volume = {34},
number = {2},
pages = {171-176},
year = {2000},
issn = {1352-2310},
doi = {https://doi.org/10.1016/S1352-2310(99)00315-5},
url = {https://www.sciencedirect.com/science/article/pii/S1352231099003155},
author = {M.W Gardner and S.R Dorling},
}

@book{jacob_introduction_1999,
	title = {Introduction to {Atmospheric} {Chemistry}},
	isbn = {978-0-691-00185-2},
	url = {http://www.jstor.org/stable/j.ctt7t8hg},
	urldate = {2024-01-09},
	publisher = {Princeton University Press},
	author = {JACOB, DANIEL J.},
	year = {1999},
}

@article{han_joint_2023,
  author={Han, Jindong and Liu, Hao and Zhu, Hengshu and Xiong, Hui},
  journal={IEEE Transactions on Knowledge and Data Engineering}, 
  title={Kill Two Birds With One Stone: A Multi-View Multi-Adversarial Learning Approach for Joint Air Quality and Weather Prediction}, 
  year={2023},
  volume={35},
  number={11},
  pages={11515-11528},
  doi={10.1109/TKDE.2023.3236423}}

@inproceedings{wang_inter-station_2021,
author = {Wang, Chunyang and Zhu, Yanmin and Zang, Tianzi and Liu, Haobing and Yu, Jiadi},
title = {Modeling Inter-Station Relationships with Attentive Temporal Graph Convolutional Network for Air Quality Prediction},
year = {2021},
isbn = {9781450382977},
publisher = {Association for Computing Machinery},
address = {New York, NY, USA},
url = {https://doi.org/10.1145/3437963.3441731},
doi = {10.1145/3437963.3441731},
booktitle = {Proceedings of the 14th ACM International Conference on Web Search and Data Mining},
pages = {616–634},
numpages = {19},
location = {Virtual Event, Israel},
series = {WSDM '21}
}

@article{yanlin_hybrid_2019,
title = {A hybrid model for spatiotemporal forecasting of PM2.5 based on graph convolutional neural network and long short-term memory},
journal = {Science of The Total Environment},
volume = {664},
pages = {1-10},
year = {2019},
issn = {0048-9697},
doi = {https://doi.org/10.1016/j.scitotenv.2019.01.333},
url = {https://www.sciencedirect.com/science/article/pii/S0048969719303821},
author = {Yanlin Qi and Qi Li and Hamed Karimian and Di Liu},
}

@inproceedings{wang_domain_knowledge_2020,
author = {Wang, Shuo and Li, Yanran and Zhang, Jiang and Meng, Qingye and Meng, Lingwei and Gao, Fei},
title = {PM2.5-GNN: A Domain Knowledge Enhanced Graph Neural Network For PM2.5 Forecasting},
year = {2020},
isbn = {9781450380195},
publisher = {Association for Computing Machinery},
address = {New York, NY, USA},
url = {https://doi.org/10.1145/3397536.3422208},
doi = {10.1145/3397536.3422208},
booktitle = {Proceedings of the 28th International Conference on Advances in Geographic Information Systems},
pages = {163–166},
numpages = {4},
location = {Seattle, WA, USA},
series = {SIGSPATIAL '20}
}

@article{demedrano_socaire_2021,
    title = {SOCAIRE: Forecasting and monitoring urban air quality in Madrid},
    journal = {Environmental Modelling \& Software},
    volume = {143},
    pages = {105084},
    year = {2021},
    issn = {1364-8152},
    doi = {https://doi.org/10.1016/j.envsoft.2021.105084},
    url = {https://www.sciencedirect.com/science/article/pii/S1364815221001274},
    author = {Rodrigo {de Medrano} and Víctor {de Buen Remiro} and José L. Aznarte},
}

@Article{brovelli_n02_2023,
    AUTHOR = {Cedeno Jimenez, Jesus Rodrigo and Pugliese Viloria, Angelly de Jesus and Brovelli, Maria Antonia},
    TITLE = {Estimating Daily NO2 Ground Level Concentrations Using Sentinel-5P and Ground Sensor Meteorological Measurements},
    JOURNAL = {ISPRS International Journal of Geo-Information},
    VOLUME = {12},
    YEAR = {2023},
    NUMBER = {3},
    ARTICLE-NUMBER = {107},
    URL = {https://www.mdpi.com/2220-9964/12/3/107},
    ISSN = {2220-9964},
    DOI = {10.3390/ijgi12030107}
}

@Article{bertrand_cams_ml_2023,
AUTHOR = {Bertrand, J.-M. and Meleux, F. and Ung, A. and Descombes, G. and Colette, A.},
TITLE = {Technical note: Improving the European air quality forecast of the Copernicus Atmosphere Monitoring Service using machine learning techniques},
JOURNAL = {Atmospheric Chemistry and Physics},
VOLUME = {23},
YEAR = {2023},
NUMBER = {9},
PAGES = {5317--5333},
URL = {https://acp.copernicus.org/articles/23/5317/2023/},
DOI = {10.5194/acp-23-5317-2023}
}

\end{document}